\documentclass[11pt]{article}
\usepackage{acl2015}
\usepackage{times}
\usepackage{latexsym}

\usepackage{amsmath, amssymb}
\usepackage{bm}
\usepackage[dvipdfmx]{graphicx}
\usepackage{url}
\usepackage{multirow}

\title{Task-Oriented Learning of Word Embeddings \\
    for Semantic Relation Classification}

\author{
	Kazuma Hashimoto$^\dagger$, Pontus Stenetorp$^\ddagger$, Makoto Miwa$^{\S}$, and Yoshimasa Tsuruoka$^\dagger$\\
	$\dagger$The University of Tokyo, 3-7-1 Hongo, Bunkyo-ku, Tokyo, Japan\\
	{\tt \{hassy,tsuruoka\}@logos.t.u-tokyo.ac.jp}\\
	$\ddagger$University College London, London, United Kingdom\\
	{\tt pontus@stenetorp.se}\\
  	$\S$Toyota Technological Institute, 2-12-1 Hisakata, Tempaku-ku, Nagoya, Japan\\
	{\tt makoto-miwa@toyota-ti.ac.jp}
}

\date{}

\begin{document}
\maketitle

\begin{abstract}
We present a novel learning method for word embeddings designed for relation classification.
Our word embeddings are trained by predicting words between noun pairs using lexical relation-specific features on a large unlabeled corpus.
This allows us to explicitly incorporate relation-specific information into the word embeddings.
The learned word embeddings are then used to construct feature vectors for a relation classification model.
On a well-established semantic relation classification task, our method significantly outperforms a baseline based on a previously introduced word embedding method, and compares favorably to previous state-of-the-art models that use syntactic information or manually constructed external resources.
\end{abstract}

%%%%%
\section{Introduction}
\begin{figure*}[t]
\begin{center}
\includegraphics[width=160mm]{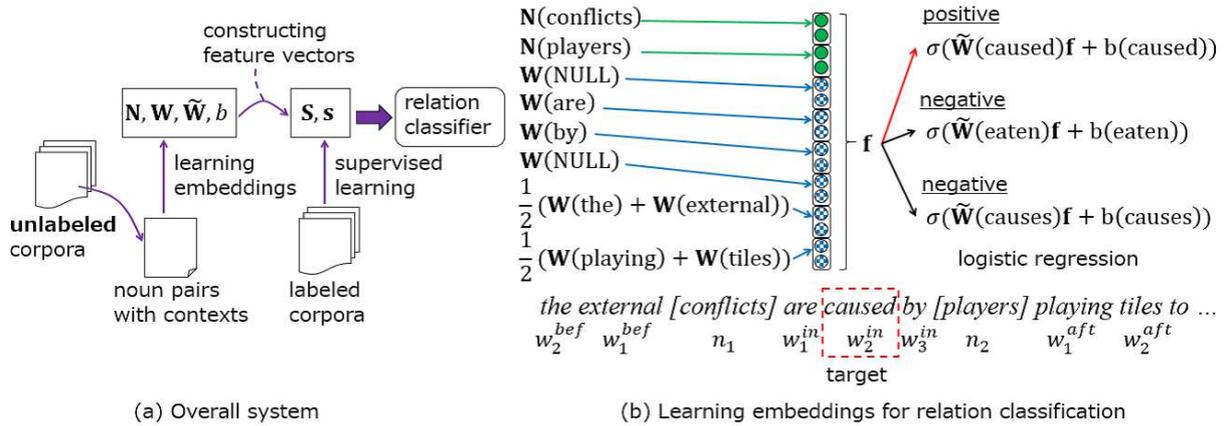}
\caption{The overview of our system (a) and the embedding learning method (b).
In the example sentence, each of {\it are}, {\it caused}, and {\it by} is treated as a target word to be predicted during training.}
\label{fig:overview}
\end{center}
\end{figure*}

Automatic classification of semantic relations has a variety of applications, such as information extraction and the construction of semantic networks~\cite{girju2007,hendrickx2010}.
A traditional approach to relation classification is to train classifiers using various kinds of features with class labels annotated by humans.
Carefully crafted features derived from lexical, syntactic, and semantic resources play a significant role in achieving high accuracy for semantic relation classification~\cite{rink2010}.

In recent years there has been an increasing interest in using {\it word embeddings} as an alternative to traditional hand-crafted features.
Word embeddings are represented as real-valued vectors and capture syntactic and semantic similarity between words.
For example, {\it word2vec}\footnote{\url{https://code.google.com/p/word2vec/}.} \cite{mikolov2013} is a well-established tool for learning word embeddings.
Although word2vec has successfully been used to learn word embeddings, these kinds of word embeddings capture only co-occurrence relationships between words~\cite{levy2014nips}.
While simply adding word embeddings trained using window-based contexts as additional features to existing systems has proven valuable~\cite{turian2010},
more recent studies have focused on how to tune and enhance word embeddings for specific tasks~\cite{bansal2014,boros2014,chen2014,guo2014,nguyen2014} and
we continue this line of research for the task of relation classification.

In this work we present a learning method for word embeddings specifically designed to be useful for relation classification.
The overview of our system and the embedding learning process are shown in Figure~\ref{fig:overview}.
First we train word embeddings by predicting each of the words between noun pairs using lexical relation-specific features on a large unlabeled corpus.
We then use the word embeddings to construct lexical feature vectors for relation classification.
Lastly, the feature vectors are used to train a relation classification model.

We evaluate our method on a well-established semantic relation classification task and compare it to a baseline based on word2vec embeddings and previous state-of-the-art models that rely on either manually crafted features, syntactic parses or external semantic resources.
Our method significantly outperforms the word2vec-based baseline, and compares favorably with previous state-of-the-art models, despite relying only on lexical level features and no external annotated resources.
Furthermore, our qualitative analysis of the learned embeddings shows that $n$-grams of our embeddings capture salient syntactic patterns similar to semantic relation types.

%%%%%
\section{Related Work}
A traditional approach to relation classification is to train classifiers in a supervised fashion using a variety of features.
These features include lexical bag-of-words features and features based on syntactic parse trees.
For syntactic parse trees, the paths between the target entities on constituency and dependency trees have been demonstrated to be useful~\cite{bunescu2005,zhang2006}.
On the shared task introduced by \newcite{hendrickx2010}, \newcite{rink2010} achieved the best score using a variety of hand-crafted features which were then used to train a Support Vector Machine (SVM).

Recently, word embeddings have become popular as an alternative to hand-crafted features~\cite{collobert2011}.
However, one of the limitations is that word embeddings are usually learned by predicting a target word in its context,
leading to only local co-occurrence information being captured~\cite{levy2014nips}.
Thus, several recent studies have focused on overcoming this limitation.
\newcite{quoc2014} integrated paragraph information into a word2vec-based model, which allowed them to capture paragraph-level information.
For dependency parsing, \newcite{bansal2014} and \newcite{chen2014} found ways to improve performance by integrating dependency-based context information into their embeddings.
\newcite{bansal2014} trained embeddings by defining parent and child nodes in dependency trees as contexts.
\newcite{chen2014} introduced the concept of feature embeddings induced by parsing a large unannotated corpus and then learning embeddings for the manually crafted features.
For information extraction, \newcite{boros2014} trained word embeddings relevant for event role extraction, and \newcite{nguyen2014} employed word embeddings for domain adaptation of relation extraction.
Another kind of task-specific word embeddings was proposed by \newcite{tang2014}, which used sentiment labels on tweets to adapt word embeddings for a sentiment analysis tasks.
However, such an approach is only feasible when a large amount of labeled data is available.

%%%%%
\section{Relation Classification Using Word Embedding-based Features}
We propose a novel method for learning word embeddings designed for relation classification.
The word embeddings are trained by {\it predicting each word between noun pairs}, given the corresponding low-level features for relation classification.
In general, to classify relations between pairs of nouns the most important features come from the pairs themselves and the words between and around the pairs~\cite{hendrickx2010}.
For example, in the sentence in Figure~\ref{fig:overview}~(b) there is a {\it cause-effect} relationship between the two nouns {\it conflicts} and {\it players}.
To classify the relation, the most common features are the noun pair ({\it conflicts, players}), the words between the noun pair ({\it are, caused, by}), the words before the pair ({\it the, external}), and the words after the pair ({\it playing, tiles, to, ...}).
As shown by \newcite{rink2010}, the words between the noun pairs are the most effective among these features.
Our main idea is to treat the most important features (the words between the noun pairs) as the targets to be predicted and other lexical features (noun pairs, words outside them) as their contexts.
Due to this, we expect our embeddings to capture relevant features for relation classification better than previous models which only use window-based contexts.

In this section we first describe the learning process for the word embeddings, focusing on lexical features for relation classification (Figure~\ref{fig:overview} (b)).
We then propose a simple and powerful technique to construct features which serve as input for a softmax classifier.
The overview of our proposed system is shown in Figure~\ref{fig:overview} (a).

\subsection{Learning Word Embeddings}
\label{subsec:learning}
Assume that there is a noun pair $\mathbf{n}=(n_1, n_2)$ in a sentence with $M_{in}$ words between the pair and $M_{out}$ words before and after the pair:
\begin{itemize}
\item $\mathbf{w}_{in}=(w^{in}_{1}, \ldots, w^{in}_{M_{in}})$~,
\item $\mathbf{w}_{bef}=(w^{bef}_{1}, \ldots, w^{bef}_{M_{out}})$~, and
\item $\mathbf{w}_{aft}=(w^{aft}_{1}, \ldots, w^{aft}_{M_{out}})$~.
\end{itemize}
Our method predicts each target word $w^{in}_{i}\in\mathbf{w}_{in}$ using three kinds of information: $\mathbf{n}$, words around $w^{in}_i$ in $\mathbf{w}_{in}$, and words in $\mathbf{w}_{bef}$ and $\mathbf{w}_{aft}$.
Words are embedded in a $d$-dimensional vector space and we refer to these vectors as word embeddings.
To discriminate between words in $\mathbf{n}$ from those in $\mathbf{w}_{in}$, $\mathbf{w}_{bef}$, and $\mathbf{w}_{aft}$, we have two sets of word embeddings: $\mathbf{N}\in\mathbb{R}^{d\times|\mathcal{N}|}$ and $\mathbf{W}\in\mathbb{R}^{d\times|\mathcal{W}|}$.
$\mathcal{W}$ is a set of words and $\mathcal{N}$ is also a set of words but contains only nouns.
Hence, the word {\it cause} has two embeddings: one in $\mathbf{N}$ and another in $\mathbf{W}$.
In general {\it cause} is used as a noun and a verb, and thus we expect the noun embeddings to capture the meanings focusing on their noun usage.
This is inspired by some recent work on word representations that explicitly assigns an independent representation for each word usage according to its part-of-speech tag~\cite{baroni2010,grefenstette2011,hashimoto2013,hashimoto2014,kartsaklis2013}.

A feature vector $\mathbf{f}\in\mathbb{R}^{2d(2+c)\times 1}$ is constructed to predict $w^{in}_{i}$ by concatenating word embeddings:
\begin{equation}
\label{eq:f}
\begin{split}
\mathbf{f}=[&\mathbf{N}(n_1); \mathbf{N}(n_2); \mathbf{W}(w^{in}_{i-1}); \ldots; \mathbf{W}(w^{in}_{i-c});\\
                 &~~~~~~~~~~~~~~~~~~~~~~~~~\mathbf{W}(w^{in}_{i+1}); \ldots; \mathbf{W}(w^{in}_{i+c});\\
                 &\frac{1}{M_{out}}\sum_{j=1}^{M_{out}}\mathbf{W}(w^{bef}_{j}); \frac{1}{M_{out}}\sum_{j=1}^{M_{out}}\mathbf{W}(w^{aft}_{j})]~.
\end{split}
\end{equation}
$\mathbf{N}(\cdot)$ and $\mathbf{W}(\cdot)\in\mathbb{R}^{d\times 1}$ corresponds to each word and $c$ is the context size.
A special {\it NULL} token is used if $i-j$ is smaller than $1$ or $i+j$ is larger than $M_{in}$ for each $j\in\{1, 2,\ldots, c\}$.

Our method then estimates a conditional probability $p(w|\mathbf{f})$ that the target word is a word $w$ given the feature vector $\mathbf{f}$, using a logistic regression model:
\begin{equation}
p(w|\mathbf{f})=\sigma(\tilde{\mathbf{W}}(w)\cdot\mathbf{f}+b(w))~,
\end{equation}
where $\tilde{\mathbf{W}}(w)\in\mathbb{R}^{2d(2+c)\times1}$ is a weight vector for $w$, $b(w)\in\mathbb{R}$ is a bias for $w$, and $\sigma(x)=\frac{1}{1+\mathrm{e}^{-x}}$ is the logistic function.
Each column vector in $\tilde{\mathbf{W}}\in\mathbb{R}^{2d(c+1)\times |\mathcal{W}|}$ corresponds to a word.
That is, we assign a logistic regression model for each word, and we can train the embeddings using the one-versus-rest approach to make $p(w^{in}_{i}|\mathbf{f})$ larger than $p(w'|\mathbf{f})$ for $w'\neq w^{in}_{i}$.
However, naively optimizing the parameters of those logistic regression models would lead to prohibitive computational cost since it grows linearly with the size of the vocabulary.

When training we employ several procedures introduced by \newcite{mikolov2013}, namely,
{\it negative sampling}, a modified unigram noise distribution and {\it subsampling}.
For negative sampling the model parameters $\mathbf{N}$, $\mathbf{W}$, $\tilde{\mathbf{W}}$, and $b$ are learned by maximizing the objective function $J_{unlabeled}$:
\begin{equation}
\label{eq:obj}
\sum_{\mathbf{n}}\sum_{i=1}^{M_{in}}\left(\log(p(w^{in}_i|\mathbf{f}))+\sum_{j=1}^{k}\log(1-p(w'_{j}|\mathbf{f}))\right)~,
\end{equation}
where $w'_{j}$ is a word randomly drawn from the unigram noise distribution weighted by an exponent of $0.75$.
Maximizing $J_{unlabeled}$ means that our method can discriminate between each target word and $k$ noise words given the target word's context.
This approach is much less computationally expensive than the one-versus-rest approach and has proven effective in learning word embeddings.

To reduce redundancy during training we use subsampling.
A training sample, whose target word is $w$, is discarded with the probability
$P_{d}(w)=1-\sqrt{\frac{t}{p(w)}}$,
where $t$ is a threshold which is set to $10^{-5}$ and $p(w)$ is a probability corresponding to the frequency of $w$ in the training corpus.
The more frequent a target word is, the more likely it is to be discarded.
To further emphasize infrequent words, we apply the subsampling approach not only to target words, but also to noun pairs;
concretely, by drawing two random numbers $r_1$ and $r_2$, a training sample whose noun pair is $(n_1, n_2)$ is discarded if $P_{d}(n_1)$ is larger than $r_1$ or $P_{d}(n_2)$ is larger than $r_2$.

Since the feature vector $\mathbf{f}$ is constructed as defined in Eq.~(\ref{eq:f}), at each training step, $\tilde{\mathbf{W}}(w)$ is updated based on information about
what pair of nouns surrounds $w$,
what word $n$-grams appear in a small window around $w$, and
what words appear outside the noun pair.
Hence, the weight vector $\tilde{\mathbf{W}}(w)$ captures rich information regarding the target word $w$.

\subsection{Constructing Feature Vectors}
\label{subsec:feature}
Once the word embeddings are trained, we can use them for relation classification.
Given a noun pair $\mathbf{n}=(n_1, n_2)$ with its context words $\mathbf{w}_{in}$, $\mathbf{w}_{bef}$, and $\mathbf{w}_{aft}$,
we construct a feature vector to classify the relation between $n_1$ and $n_2$ by concatenating three kinds of feature vectors:
\begin{description}
\item[$\mathbf{g}_\mathrm{n}$] the word embeddings of the noun pair,
\item[$\mathbf{g}_\mathrm{in}$] the averaged $n$-gram embeddings between the pair, and
\item[$\mathbf{g}_\mathrm{out}$] the concatenation of the averaged word embeddings in $\mathbf{w}_{bef}$ and $\mathbf{w}_{aft}$.
\end{description}

%The noun pair itself is a strong indicator to classify the relation.
The feature vector $\mathbf{g}_\mathrm{n}\in\mathbb{R}^{2d\times 1}$ is the concatenation of $\mathbf{N}(n_1)$ and $\mathbf{N}(n_2)$:
\begin{equation}
\mathbf{g}_\mathrm{n}=[\mathbf{N}(n_1); \mathbf{N}(n_2)]~.
\end{equation}

Words between the noun pair contribute to classifying the relation, and
one of the most common ways to incorporate an arbitrary number of words is treating them as a bag of words.
However, word order information is lost for bag-of-words features such as averaged word embeddings.
To incorporate the word order information, we first define $n$-gram embeddings $\mathbf{h}_{i}\in\mathbb{R}^{4d(1+c)\times 1}$ between the noun pair:
\begin{equation}
\label{eq:n-gram-base}
\begin{split}
\mathbf{h}_{i}=[&\mathbf{W}(w^{in}_{i-1}); \ldots; \mathbf{W}(w^{in}_{i-c});\\
                    &\mathbf{W}(w^{in}_{i+1}); \ldots; \mathbf{W}(w^{in}_{i+c}); \tilde{\mathbf{W}}(w^{in}_{i})]~.
\end{split}
\end{equation}
Note that $\tilde{\mathbf{W}}$ can also be used and that the value used for $n$ is $(2c+1)$.
As described in Section~\ref{subsec:learning}, $\tilde{\mathbf{W}}$ captures meaningful information about each word and after the first embedding learning step we can treat the embeddings in $\tilde{\mathbf{W}}$ as features for the words.
\newcite{mnih2013} have demonstrated that using embeddings like those in $\tilde{\mathbf{W}}$ is useful in representing the words.
We then compute the feature vector $\mathbf{g}_\mathrm{in}$ by averaging $\mathbf{h}_i$:
\begin{equation}
\mathbf{g}_\mathrm{in}=\frac{1}{M_{in}}\sum_{i=1}^{M_{in}}\mathbf{h}_{i}~.
\end{equation}
We use the averaging approach since $M_{in}$ depends on each instance.
The feature vector $\mathbf{g}_\mathrm{in}$ allows us to represent word sequences of arbitrary lengths as fixed-length feature vectors using the simple operations: concatenation and averaging.

The words before and after the noun pair are sometimes important in classifying the relation.
For example, in the phrase ``pour $n_1$ into $n_2$'', the word {\it pour} should be helpful in classifying the relation.
As with Eq.~(\ref{eq:f}), we use the concatenation of the averaged word embeddings of words before and after the noun pair to compute the feature vector $\mathbf{g}_\mathrm{out}\in\mathbb{R}^{2d\times 1}$:
\begin{equation}
\mathbf{g}_\mathrm{out}=\frac{1}{M_{out}}[\sum_{j=1}^{M_{out}}\mathbf{W}(w^{bef}_{j});\sum_{j=1}^{M_{out}}\mathbf{W}(w^{aft}_{j})]~.
\end{equation}

As described above, the overall feature vector $\mathbf{e}\in\mathbb{R}^{4d(2+c)\times 1}$ is constructed by concatenating $\mathbf{g}_\mathrm{n}$, $\mathbf{g}_\mathrm{in}$, and $\mathbf{g}_\mathrm{out}$.
We would like to emphasize that we only use simple operations: averaging and concatenating the learned word embeddings.
The feature vector $\mathbf{e}$ is then used as input for a softmax classifier, without any complex transformation such as matrix multiplication with non-linear functions.

\subsection{Supervised Learning}
\label{subsec:sup}
Given a relation classification task we train a softmax classifier using the feature vector $\mathbf{e}$ described in Section~\ref{subsec:feature}.
For each $k$-th training sample with a corresponding label $l_k$ among $L$ predefined labels, we compute a conditional probability given its feature vector $\mathbf{e}_k$:
\begin{equation}
p(l_k|\mathbf{e}_k)=\frac{\exp(\mathbf{o}(l_k))}{\sum_{i=1}^{L}\exp(\mathbf{o}(i))},
\end{equation}
where $\mathbf{o}\in\mathbb{R}^{L\times 1}$ is defined as $\mathbf{o}=\mathbf{S}\mathbf{e}_k+\mathbf{s}$, and $\mathbf{S}\in\mathbb{R}^{L\times 4d(2+c)}$ and $\mathbf{s}\in\mathbb{R}^{L\times 1}$ are the softmax parameters.
$\mathbf{o}(i)$ is the $i$-th element of $\mathbf{o}$.
We then define the objective function as:
\begin{equation}
J_{labeled}=\sum_{k=1}^{K}\log(p(l_k|\mathbf{e}_k))-\frac{\lambda}{2}\|\mathbf{\theta}\|^2~.
\end{equation}
$K$ is the number of training samples and $\lambda$ controls the L-2 regularization.
$\theta=(\mathbf{N}, \mathbf{W}, \tilde{\mathbf{W}}, \mathbf{S}, \mathbf{s})$ is the set of parameters and $J_{labeled}$ is maximized using AdaGrad~\cite{duchi2011}.
We have found that {\it dropout}~\cite{hinton2012} is helpful in preventing our model from overfitting.
Concretely, elements in $\mathbf{e}$ are randomly omitted with a probability of $0.5$ at each training step.
Recently dropout has been applied to deep neural network models for natural language processing tasks and proven effective~\cite{dropout1,dropout2}.

In what follows, we refer to the above method as \textbf{RelEmb}.
While RelEmb uses only low-level features, a variety of useful features have been proposed for relation classification.
Among them, we use dependency path features~\cite{bunescu2005} based on the untyped binary dependencies of the Stanford parser to find the shortest path between target nouns.
The dependency path features are computed by averaging word embeddings from $\mathbf{W}$ on the shortest path, and are then concatenated to the feature vector $\mathbf{e}$.
Furthermore, we directly incorporate semantic information using word-level semantic features from Named Entity (NE) tags and WordNet hypernyms, as used in previous work~\cite{rink2010,socher2012,moyu2014}.
We refer to this extended method as \textbf{RelEmb$_\mathrm{FULL}$}.
Concretely, RelEmb$_\mathrm{FULL}$ uses the same binary features as in \newcite{socher2012}.
The features come from NE tags and WordNet hypernym tags of target nouns provided by a sense tagger~\cite{ciaramita2006}.

%%%%%
\section{Experimental Settings}

\subsection{Training Data}
\label{subsec:data}
For pre-training we used a snapshot of the English Wikipedia\footnote{\url{http://dumps.wikimedia.org/enwiki/}.} from November 2013.
First, we extracted 80 million sentences from the original Wikipedia file, and then used
{\it Enju}\footnote{
Despite Enju being a syntactic parser we only use the POS tagger component.
The accuracy of the POS tagger is about 97.2$\%$ on the WSJ corpus.
%There are other taggers which are significantly faster, for example spaCy, while maintaining the same level of accuracy.
}~\cite{miyao2008}
to automatically assign part-of-speech (POS) tags.
From the POS tags we used {\it NN}, {\it NNS}, {\it NNP}, or {\it NNPS} to locate noun pairs in the corpus.
We then collected training data by listing pairs of nouns and the words between, before, and after the noun pairs.
A noun pair was omitted if the number of words between the pair was larger than 10 and 
we consequently collected 1.4 billion pairs of nouns and their contexts
\footnote{The training data, the training code, and the learned model parameters used in this paper are publicly
available at \url{http://www.logos.t.u-tokyo.ac.jp/~hassy/publications/conll2015/}}.
We used the 300,000 most frequent words and the 300,000 most frequent nouns and treated out-of-vocabulary words as a special {\it UNK} token.

\subsection{Initialization and Optimization}
\label{subsec:init}
We initialized the embedding matrices $\mathbf{N}$ and $\mathbf{W}$ with zero-mean gaussian noise with a variance of $\frac{1}{d}$.
$\tilde{\mathbf{W}}$ and $b$ were zero-initialized.
The model parameters were optimized by maximizing the objective function in Eq.~(\ref{eq:obj}) using stochastic gradient ascent.
The learning rate was set to $\alpha$ and linearly decreased to $0$ during training, as described in \newcite{mikolov2013b}.
The hyperparameters are the embedding dimensionality $d$, the context size $c$, the number of negative samples $k$, the initial learning rate $\alpha$, and $M_{out}$, the number of words outside the noun pairs.
For hyperparameter tuning, we first fixed $\alpha$ to $0.025$ and $M_{out}$ to $5$, and then set $d$ to $\{50,~100,~300\}$, $c$ to $\{1,~2,~3\}$, and $k$ to $\{5,~15,~25\}$.

At the supervised learning step, we initialized $\mathbf{S}$ and $\mathbf{s}$ with zeros.
The hyperparameters, the learning rate for AdaGrad, $\lambda$, $M_{out}$, and the number of iterations, were determined via 10-fold cross validation on the training set for each setting.
Note that $M_{out}$ can be tuned at the supervised learning step, adapting to a specific dataset.

%%%%%
\section{Evaluation}

\subsection{Evaluation Dataset}
\label{subsec:eval_data}
We evaluated our method on the SemEval 2010 Task 8 data set\footnote{\url{http://docs.google.com/View?docid=dfvxd49s_36c28v9pmw}.} \cite{hendrickx2010}, which involves predicting the semantic relations between noun pairs in their contexts.
The dataset, containing 8,000 training and 2,717 test samples, defines nine classes ({\it Cause-Effect, Entity-Origin, etc.}) for ordered relations and one class ({\it Other}) for other relations.
Thus, the task can be treated as a 19-class classification task.
Two examples from the training set are shown below.
\begin{itemize}
\item[(a)] Financial [stress]$_{\mathrm{E_1}}$ is one of the main causes of [divorce]$_{\mathrm{E_2}}$
\item[(b)] The [burst]$_{\mathrm{E_1}}$ has been caused by water hammer [pressure]$_{\mathrm{E_2}}$
\end{itemize}
Training example (a) is classified as {\it Cause-Effect(E$_1$, E$_2$)} which denotes that {\it E$_2$} is an effect caused by {\it E$_1$},
while training example (b) is classified as {\it Cause-Effect(E$_2$, E$_1$)} which is the inverse of {\it Cause-Effect(E$_1$, E$_2$)}.
We report the official macro-averaged F1 scores and accuracy.

\subsection{Models}
To empirically investigate the performance of our proposed method we compared
it to several baselines and previously proposed models.

\subsubsection{Random and word2vec Initialization}
\paragraph{Rand-Init.}
The first baseline is RelEmb itself, but without applying the learning method on the unlabeled corpus.
In other words, we train the softmax classifier from Section~\ref{subsec:sup} on the labeled training data with randomly initialized model parameters.

\paragraph{W2V-Init.}
The second baseline is RelEmb using word embeddings learned by word2vec.
More specifically, we initialize the embedding matrices $\mathbf{N}$ and $\mathbf{W}$ with the word2vec embeddings.
Related to our method, word2vec has a set of weight vectors similar to $\tilde{\mathbf{W}}$ when trained with negative sampling and we use these weight vectors as a replacement for $\tilde{\mathbf{W}}$.
We trained the word2vec embeddings using the {\it CBOW} model with subsampling on the full Wikipedia corpus.
As with our experimental settings, we fix the learning rate to $0.025$, and investigate several hyperparameter settings.
For hyperparameter tuning we set the embedding dimensionality $d$ to $\{50,~100,~300\}$, the context size $c$ to $\{1,~3,~9\}$, and the number of negative samples $k$ to $\{5,~15,~25\}$.

\subsubsection{SVM-Based Systems}
A simple approach to the relation classification task is to use SVMs with standard binary bag-of-words features.
The bag-of-words features included the noun pairs and words between, before, and after the pairs, and we
used LIBLINEAR\footnote{\url{http://www.csie.ntu.edu.tw/~cjlin/liblinear/}.} as our classifier.

\subsubsection{Neural Network Models}
\newcite{socher2012} used Recursive Neural Network (RNN) models to classify the relations.
Subsequently, \newcite{crnn2015} and \newcite{hashimoto2013} proposed RNN models to better handle the relations.
These methods rely on syntactic parse trees.

\newcite{moyu2014} introduced their novel Factor-based Compositional Model (FCM) and presented results from several model variants,
the best performing being FCM$_\mathrm{EMB}$ and FCM$_\mathrm{FULL}$.
The former only uses word embedding information and the latter relies on dependency paths and NE features,
in addition to word embeddings.

\newcite{zeng2014} used a Convolutional Neural Network (CNN) with WordNet hypernyms.
Noteworthy in relation to the RNN-based methods, the CNN model does not rely on parse trees.
More recently, \newcite{dos2015} have introduced CR-CNN  by extending the CNN model and achieved the best result to date.
The key point of CR-CNN is that it improves the classification score by omitting the noisy class ``Other'' in the dataset described in Section~\ref{subsec:eval_data}.
We call CR-CNN using the ``Other'' class CR-CNN$_\mathrm{Other}$ and CR-CNN omitting the class CR-CNN$_\mathrm{Best}$.

\subsection{Results and Discussion}

\begin{table*}[t]
\begin{center}
%{\small
{\fontsize{10.2pt}{12pt}\selectfont
\begin{tabular}{l|l|c}
& Features for classifiers & F1 / ACC ($\%$)\\\hline\hline
RelEmb$_\mathrm{FULL}$ & embeddings, dependency paths, WordNet, NE & 83.5 / 79.9 \\
RelEmb & embeddings & 82.8 / 78.9 \\\hline
RelEmb (W2V-Init) & embeddings & 81.8 / 77.7 \\
RelEmb (Rand-Init) & embeddings & 78.2 / 73.5 \\
SVM & bag of words & 76.5 / 72.0 \\\hline
SVM & bag of words, POS, dependency paths, WordNet, & \multirow{2}{*}{82.2 / 77.9} \\
\cite{rink2010} & paraphrases, TextRunner, Google $n$-grams, etc. & \\\hline
CR-CNN$_\mathrm{Best}$~\cite{dos2015} & embeddings, word position embeddings & 84.1 / n/a \\
FCM$_\mathrm{FULL}$~\cite{moyu2014} & embeddings, dependency paths, NE & 83.0 / n/a \\
CR-CNN$_\mathrm{Other}$~\cite{dos2015} & embeddings, word position embeddings & 82.7 / n/a \\
CRNN~\cite{crnn2015} & embeddings, parse trees, WordNet, NE, POS & 82.7 / n/a \\
CNN~\cite{zeng2014} & embeddings, WordNet & 82.7 / n/a \\
MVRNN~\cite{socher2012} & embeddings, parse trees, WordNet, NE, POS & 82.4 / n/a \\
FCM$_\mathrm{EMB}$~\cite{moyu2014} & embeddings & 80.6 / n/a \\
RNN~\cite{hashimoto2013} & embeddings, parse trees, phrase categories, etc. & 79.4 / n/a\\\hline
\end{tabular}
}
\caption{Scores on the test set for SemEval 2010 Task 8.}
\label{tb:test}
\end{center}
\end{table*}

The scores on the test set for SemEval 2010 Task 8 are shown in Table~\ref{tb:test}.
RelEmb achieves 82.8$\%$ of F1 which is better than those of almost all models compared and comparable to that of the previous state of the art, except for CR-CNN$_\mathrm{Best}$.
Note that RelEmb does not rely on external semantic features and syntactic parse features\footnote{While we use a POS tagger to locate noun pairs, RelEmb does not explicitly use POS features at the supervised learning step.}.
Furthermore, RelEmb$_\mathrm{FULL}$ achieves 83.5$\%$ of F1.
We calculated a confidence interval $(82.0,~84.9)$ ($p<0.05$) using bootstrap resampling~\cite{bootstrap}.

\subsubsection{Comparison with the Baselines}
RelEmb significantly outperforms not only the Rand-Init baseline, but also the W2V-Init baseline.
These results show that our task-specific word embeddings are more useful than those trained using window-based contexts.
A point that we would like to emphasize is that the baselines are unexpectedly strong.
As was noted by \newcite{wang2012}, we should carefully implement strong baselines and see whether complex models can outperform these baselines.

\subsubsection{Comparison with SVM-Based Systems}
RelEmb performs much better than the bag-of-words-based SVM.
This is not surprising given that we use a large unannotated corpus and embeddings with a large number of parameters.
RelEmb also outperforms the SVM system of \newcite{rink2010}, which demonstrates the effectiveness of our task-specific word embeddings, despite our only requirement being a large unannotated corpus and a POS tagger.

\subsubsection{Comparison with Neural Network Models}
RelEmb outperforms the RNN models.
In our preliminary experiments, we have found some undesirable parse trees when computing vector representations using RNN-based models and
such parsing errors might hamper the performance of the RNN models.

FCM$_\mathrm{FULL}$, which relies on dependency paths and NE features, achieves a better score than that of RElEmb.
Without such features, RelEmb outperforms FCM$_\mathrm{EMB}$ by a large margin.
By incorporating external resources, RelEmb$_\mathrm{FULL}$ outperforms FCM$_\mathrm{FULL}$.

RelEmb compares favorably to CR-CNN$_\mathrm{Other}$, despite our method being less computationally expensive than CR-CNN$_\mathrm{Other}$.
When classifying an instance, the number of the floating number multiplications is $4d(2+c)L$ in our method since our method requires only one matrix-vector product for the softmax classifier as described in Section~\ref{subsec:sup}.
$c$ is the window size, $d$ is the word embedding dimensionality, and $L$ is the number of the classes.
In CR-CNN$_\mathrm{Other}$, the number is $(Dc(d+2d')N+DL)$, where $D$ is the dimensionality of the convolution layer, $d'$ is the position embedding dimensionality, and $N$ is the average length of the input sentences.
Here, we omit the cost of the hyperbolic tangent function in CR-CNN$_\mathrm{Other}$ for simplicity.
Using the best hyperparameter settings, the number is roughly $3.8\times 10^4$ in our method, and $1.6\times 10^7$ in CR-CNN$_\mathrm{Other}$ assuming $N$ is 10.
\newcite{dos2015} also boosted the score of CR-CNN$_\mathrm{Other}$ by omitting the noisy class ``Other'' by a ranking-based classifier, and achieved the best score (CR-CNN$_\mathrm{Best}$).
Our results may also be improved by using the same technique, but the technique is dataset-dependent, so we did not incorporate the technique.

\subsection{Analysis on Training Settings}

\begin{table}[t]

\begin{center}
%{\small
\begin{tabular}{r|r||c|c|c}
$c$ & $d$ & $k=5$ & $k=15$ & $k=25$ \\ \hline \hline
\multirow{2}{*}{$1$} & 50  & 80.5 & 81.0 & 80.9 \\
                           & 100 & 80.9 & 81.3 & 81.2 \\ \hline
\multirow{2}{*}{$2$} & 50  & 80.9 & 81.3 & 81.3 \\
                           & 100 & 81.3 & 81.6 & 81.7 \\ \hline
\multirow{3}{*}{$3$} & 50  & 81.0 & 81.0 & 81.5 \\
                           & 100 & 81.3 & 81.9 & \textbf{82.2} \\
                           & 300 & - & - & 82.0 \\ \hline
\end{tabular}
%}
\caption{Cross-validation results for RelEmb.}
\label{tb:cross_proposed}
\end{center}
\end{table}

\begin{table}[t]
\begin{center}
%{\small
\begin{tabular}{r|r||c|c|c}
$c$ & $d$ & $k=5$ & $k=15$ & $k=25$ \\ \hline \hline
\multirow{3}{*}{$1$} & 50  & 80.5 & 80.7 & 80.9 \\
                           & 100 & 81.1 & 81.2 & 81.0 \\
                           & 300 & 81.2 & \textbf{81.3} & 81.2 \\ \hline
\multirow{2}{*}{$3$} & 50  & 80.4 & 80.7 & 80.8 \\
                           & 100 & 81.0 & 81.0 & 80.9 \\ \hline
\multirow{2}{*}{$9$} & 50  & 80.0 & 79.8 & 80.2 \\
                           & 100 & 80.3 & 80.4 & 80.1 \\ \hline
\end{tabular}
%}
\caption{Cross-validation results for the W2V-Init.}
\label{tb:cross_word2vec}
\end{center}
\end{table}

\begin{table}[t]
\begin{center}
%{\small
\begin{tabular}{ccccc}
$\mathbf{g}_\mathrm{n}$ &
$\mathbf{g}_\mathrm{in}$ &
$\mathbf{g}_\mathrm{in}'$ &
$\mathbf{g}_\mathrm{n}, \mathbf{g}_\mathrm{in}$ &
$\mathbf{g}_\mathrm{n}, \mathbf{g}_\mathrm{in}, \mathbf{g}_\mathrm{out}$\\\hline\hline
61.8 & 70.2 & 68.2 & 81.1 & 82.2 \\\hline
\end{tabular}
%}
\caption{Cross-validation results for ablation tests.}
\label{tb:ablation}
\end{center}
\end{table}

We perform analysis of the training procedure focusing on RelEmb.

\subsubsection{Effects of Tuning Hyperparameters}

In Tables~\ref{tb:cross_proposed} and \ref{tb:cross_word2vec}, we show how tuning the hyperparameters of our method and word2vec affects the classification results using 10-fold cross validation on the training set.
The same split is used for each setting, so all results are comparable to each other.
The best settings for the cross validation are used to produce the results reported in Table~\ref{tb:test}.

Table~\ref{tb:cross_proposed} shows F1 scores obtained by RelEmb.
The results for $d=50,~100$ show that RelEmb benefits from relatively large context sizes.
The $n$-gram embeddings in RelEmb capture richer information by setting $c$ to 3 compared to setting $c$ to 1.
Relatively large numbers of negative samples also slightly boost the scores.
As opposed to these trends, the score does not improve using $d=300$.
We use the best setting ($c=3$, $d=100$, $k=25$) for the remaining analysis.
We note that RelEmb$_\mathrm{FULL}$ achieves an F1-score of 82.5.

We also performed similar experiments for the W2V-Init baseline, and the results are shown in Table~\ref{tb:cross_word2vec}.
In this case, the number of negative samples does not affect the scores, and the best score is achieved by $c=1$.
As discussed in \newcite{bansal2014}, the small context size captures the syntactic similarity between words rather than the topical similarity.
This result indicates that syntactic similarity is more important than topical similarity for this task.
Compared to the word2vec embeddings, our embeddings capture not only local context information using word order, but also long-range co-occurrence information by being tailored for the specific task.

\subsubsection{Ablation Tests}

\begin{table}[t]
\begin{center}
%{\small
\begin{tabular}{l|c}
Method & Score \\ \hline \hline
RelEmb $\mathbf{N}$ & 0.690 \\
RelEmb $\mathbf{W}$ & 0.599 \\ \hline
W2V-Init & 0.687 \\ \hline
\end{tabular}
%}
\caption{Evaluation on the WordSim-353 dataset.}
\label{tb:wordsim}
\end{center}
\end{table}

\begin{table*}[ht]
\begin{center}
{\fontsize{7.8pt}{9pt}\selectfont
\begin{tabular}{l|lll||l|p{2.5em}lp{2.5em}||l|p{3.5em}p{3em}p{2.5em}}
\multicolumn{4}{c}{\textbf{Cause-Effect(E$_1$,E$_2$)}} & \multicolumn{4}{c}{\textbf{Content-Container(E$_1$,E$_2$)}} & \multicolumn{4}{c}{\textbf{Message-Topic(E$_1$,E$_2$)}}\\
resulted & poverty & caused & the & inside & was & inside & a & discuss & magazines & relating & to\\
caused & stability & caused & the & in & was & in & a & explaining & to & discuss & aspects \\
generated & coast & resulted & in & hidden & hidden & in & a & discussing & concerned & about & NULL\\
cause & fire & caused & due & was & was & inside & the & relating & interview & relates & to \\
causes & that & resulted & in & stored & was & hidden & in & describing & to & discuss & the \\\hline
\multicolumn{12}{c}{}\\
\multicolumn{4}{c}{\textbf{Cause-Effect(E$_2$,E$_1$)}} & \multicolumn{4}{c}{\textbf{Content-Container(E$_2$,E$_1$)}} & \multicolumn{4}{c}{\textbf{Message-Topic(E$_2$,E$_1$)}}\\
after & caused & by & radiation & full & NULL & full & of & subject & were & related & in\\
from & caused & by & infection & included & was & full & of & related & was & related & in\\
caused & stomach & caused & by & contains & a & full & NULL & discussed &  been & discussed & in\\
triggered & caused & by & genetic & contained & a & full & and & documented & is & related & through\\
due & anger & caused & by & stored & a & full & forty & received & the & subject & of \\\hline
\end{tabular}
}
\caption{Top five unigrams and trigrams with the highest scores for six classes.}
\label{tb:n-gram}
\end{center}
\end{table*}

As described in Section~\ref{subsec:feature}, we concatenate three kinds of feature vectors, $\mathbf{g}_\mathrm{n}$, $\mathbf{g}_\mathrm{in}$, and $\mathbf{g}_\mathrm{out}$, for supervised learning.
Table~\ref{tb:ablation} shows classification scores for ablation tests using 10-fold cross validation.
We also provide a score using a simplified version of $\mathbf{g}_\mathrm{in}$, where the feature vector $\mathbf{g}_\mathrm{in}'$ is computed by averaging the word embeddings $[\mathbf{W}(w_{i}^{in}); \tilde{\mathbf{W}}(w_{i}^{in})]$ of the words between the noun pairs.
This feature vector $\mathbf{g}_\mathrm{in}'$ then serves as a bag-of-words feature.

Table~\ref{tb:ablation} clearly shows that the averaged $n$-gram embeddings contribute the most to the semantic relation classification performance.
The difference between the scores of $\mathbf{g}_\mathrm{in}$ and $\mathbf{g}_\mathrm{in}'$ shows the effectiveness of our averaged $n$-gram embeddings.

\subsubsection{Effects of Dropout}
At the supervised learning step we use dropout to regularize our model.
Without dropout, our performance drops from 82.2$\%$ to 81.3$\%$ of F1 on the training set using 10-fold cross validation.

\subsubsection{Performance on a Word Similarity Task}
As described in Section~\ref{subsec:learning}, we have the noun-specific embeddings $\mathbf{N}$ as well as the standard word embeddings $\mathbf{W}$.
We evaluated the learned embeddings using a word-level semantic evaluation task called {\it WordSim-353}~\cite{wordsim353}.
This dataset consists of 353 pairs of nouns and each pair has an averaged human rating which corresponds to a semantic similarity score.
Evaluation is performed by measuring Spearman's rank correlation between the human ratings and the cosine similarity scores of the embeddings.
Table~\ref{tb:wordsim} shows the evaluation results.
We used the best settings reported in Table~\ref{tb:cross_proposed} and \ref{tb:cross_word2vec} since our method is designed for relation classification and it is not clear how to tune the hyperparameters for the word similarity task.
As shown in the result table, the noun-specific embeddings perform better than the standard embeddings in our method, which indicates the noun-specific embeddings capture more useful information in measuring the semantic similarity between nouns.
The performance of the noun-specific embeddings is roughly the same as that of the word2vec embeddings.

%%%%%
\subsection{Qualitative Analysis on the Embeddings}
\label{subsec:quali-n}

Using the $n$-gram embeddings $\mathbf{h}_i$ in Eq.~(\ref{eq:n-gram-base}), we inspect which $n$-grams are relevant to each relation class after the supervised learning step of RelEmb.
When the context size $c$ is $3$, we can use at most $7$-grams.
The learned weight matrix $\mathbf{S}$ in Section~\ref{subsec:sup} is used to detect the most relevant $n$-grams for each class.
More specifically, for each $n$-gram embedding $(n=1,3)$ in the training set, we compute the dot product between the $n$-gram embedding and the corresponding components in $\mathbf{S}$.
We then select the pairs of $n$-grams and class labels with the highest scores.
In Table~\ref{tb:n-gram} we show the top five $n$-grams for six classes.
These results clearly show that the $n$-gram embeddings capture salient syntactic patterns which are useful for the relation classification task.

%%%%%
\section{Conclusions and Future Work}
We have presented a method for learning word embeddings specifically designed for relation classification.
The word embeddings are trained using large unlabeled corpora to capture lexical features for relation classification.
On a well-established semantic relation classification task our method significantly outperforms the baseline based on word2vec.
Our method also compares favorably to previous state-of-the-art models that rely on syntactic parsers and external semantic resources, despite our method requiring only access to an unannotated corpus and a POS tagger.
For future work, we will investigate how well our method performs on other domains and datasets and how relation labels can help when learning embeddings in a semi-supervised learning setting.

\section*{Acknowledgments}
We thank the anonymous reviewers for their helpful comments and suggestions.
%This work was supported by JSPS KAKENHI Grant Number 13F03041.

\bibliographystyle{acl}
\bibliography{bibtex}

\end{document}